\def\year{2021}\relax
\documentclass[letterpaper]{article}
\usepackage{aaai21}
\usepackage{times}
\usepackage{helvet}
\usepackage{courier}
\usepackage[hyphens]{url}
\usepackage{graphicx}
\usepackage{natbib}
\usepackage{caption}
\usepackage[utf8]{inputenc}
\usepackage[T1]{fontenc}
\usepackage{hyperref}
\usepackage{amsmath}
\usepackage{amsfonts}
\usepackage{bm}
\usepackage{xfrac}
\usepackage{xspace}
\usepackage{lmodern}

\nocopyright
\graphicspath{{figures/}}

\renewcommand*{\r}{\bm{r}}
\newcommand*{\x}{\bm{x}}
\newcommand*{\w}{\bm{w}}
\newcommand*{\z}{\bm{z}}

\newcommand*{\p}{\bm{\phi}}
\newcommand*{\0}{\bm{0}}
\newcommand*{\1}{\bm{1}}
\newcommand*{\s}{\bm{s}}
\newcommand{\TXP}{\texttt{timeXplain}\xspace}

\DeclareMathOperator*{\argmin}{arg\,min}
\DeclareMathOperator*{\RDFT}{rDFT}
\DeclareMathOperator*{\iRDFT}{(rDFT)^{-1}}
\DeclareMathOperator*{\Shap}{\Phi}
\DeclareMathOperator*{\R}{R}

\begingroup
\makeatletter
\@for\theoremstyle:=definition,remark,plain\do{%
	\expandafter\g@addto@macro\csname th@\theoremstyle\endcsname{%
		\addtolength\thm@preskip\parskip
	}%
}
\endgroup
\newtheorem{definition}{Definition}

\title{timeXplain -- A Framework for Explaining the\\Predictions of Time Series Classifiers}

\author{
	Felix Mujkanovic\hspace{0.05em}\textsuperscript{\rm 1}, Vanja Dosko\v{c}\hspace{0.1em}\textsuperscript{\rm 2}, Martin Schirneck\hspace{0.05em}\textsuperscript{\rm 3},\\
	Patrick Sch\"afer\hspace{0.05em}\textsuperscript{\rm 4}, Tobias Friedrich\hspace{0.05em}\textsuperscript{\rm 2}\\
}
\affiliations{
	\textsuperscript{\rm 1}Max Planck Institute for Informatics, Saarbr\"ucken, Germany\\
	\textsuperscript{\rm 2}Hasso Plattner Institute, University of Potsdam, Germany\\
	\textsuperscript{\rm 3}Faculty of Computer Science, University of Vienna, Austria\\
	\textsuperscript{\rm 4}Department of Computer Science, Humboldt University of Berlin, Germany\\
	felix.mujkanovic@mpi-inf.mpg.de\\
}

\begin{document}

\maketitle

\begin{abstract}
Modern time series classifiers display impressive predictive capabilities, yet their decision-making processes mostly remain black boxes to the user. At the same time, model-agnostic explainers, such as the recently proposed \texttt{SHAP}, promise to make the predictions of machine learning models interpretable, provided there are well-designed domain mappings. We bring both worlds together in our \TXP framework, extending the reach of explainable artificial intelligence to time series classification and value prediction. We present novel domain mappings for the time domain, frequency domain, and time series statistics and analyze their explicative power as well as their limits. We employ a novel evaluation metric to experimentally compare \TXP to several model-specific explanation approaches for state-of-the-art time series classifiers.\footnote{Code: \url{https://github.com/LoadingByte/timeXplain}}
\end{abstract}

\section{Introduction}

Artificial intelligence technology has become ubiquitous in industry and our everyday life in recent years. From dynamic pricing and recommender systems to autonomous vehicles and predictive policing, machine learning (ML) models continuously expand their influence on society. As they are entrusted with more and more responsibility, we should wonder whether we can actually trust them. ML systems must be safe, exhibit predictable and expected behavior, and must not discriminate~\cite{lipton16}. Whether a system satisfies such fuzzy, yet indispensable criteria cannot be determined by computing a single ``trustworthiness'' metric. Instead, researchers are pushing towards technologies that \emph{explain} the decision-making process of ML models. Through understanding the reasoning behind the predictions of complicated estimators, engineers are enabled to both spot flaws and build trust in their systems. The movement towards explainable artificial intelligence has been further fueled by a ``right to explanation'' recently proposed by the European Union~\cite{goodman17}. The Communications of the ACM dedicated their January 2020 issue to interpretable machine learning. In the cover article, \citet{Du20Communications} categorized the existing approaches by two major criteria: are the techniques intrinsic or post-hoc and are they global or local. Intrinsic methods are specialized ML models that have interpretability built into them~\cite{alvarez18selfexpl}. Post-hoc approaches treat the model as a black box. Also, the derived explanations can either be global, aiming at understanding the whole model, or they can be local, focusing on the prediction with respect to a specific input. All the categories have in common that one's ability to comprehend the decision-making process highly depends on the model complexity. Understanding a linear regression might be feasible~\cite{lou12}, but grasping the whole inner workings of a neural network with millions of weights is considered impossible today~\cite{lipton16}.

\begin{figure}[t]
	\vspace*{1em}
	\centering
	\includegraphics[width=\columnwidth]{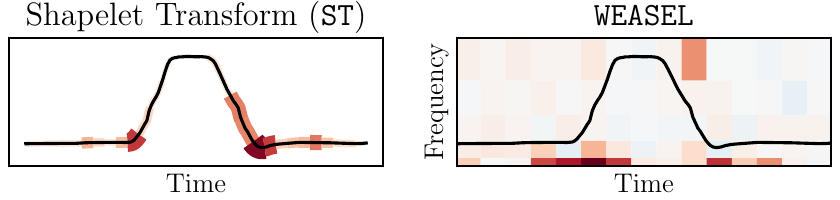}
	\caption{\TXP explains how the \texttt{ST} and \texttt{WEASEL} classifiers predict time series \#2 from the UCR GunPoint test set. The colors show which time slices (for \texttt{ST}) or frequency bands (for \texttt{WEASEL}) contribute towards (red) or against (blue) the correct classification as ``point''. Confirming the previous literature, the \texttt{ST} classifier mostly focuses on the dip after the actor lowered their arm~\cite{hills13shapeletJournal}. \texttt{WEASEL} focuses on low frequencies during the draw and dip, and high frequencies during lowering, presumably the small bump.}
	\label{fig:exemplary expl}
\end{figure}

\textit{Time series} are sequences of values whose fluctuations are associated with the passing of time. All sensor data is given this way and the pervasive use of monitoring devices in digital health~\cite{Rajkomar18health}, predictive maintenance~\cite{Lin19predictiveMaintenance}, or cyber-security~\cite{Susto18cyber} has led to an enormous increase of available time series data. Automated time series estimators (TSE) are applied to process this data. This subsumes all machine learning models that assign any kind of number to a time series, including classification~\cite{bagnall17,fawaz19,Schafer16Scalable} and other value prediction tasks. With the rise of machine learning applications on time series, explanations of their behavior are much needed. However, existing local approaches, like class activation maps for neural networks~\cite{fawaz19} and superposition of shapelets~\cite{hills13shapeletJournal} or symbolic aggregate/Fourier approximation windows~\cite{nguyen19InterpretableLinearJournal} are limited to few models. To enable the explanation of any model, we propose \TXP, a post-hoc local explanation framework for time series estimators.

To generate a local explanation, \emph{model-agnostic} approaches often use perturbation in that they query the ML model in the vicinity of the input whose analysis they want to explain. We call this input the \emph{specimen}. By slightly altering its features, one can gain insight into which of those features are relevant to the model, yielding an \emph{explanation} as the result. To flesh out such an approach, one needs to define \emph{mapping functions} detailing how to perturb the specimen. \citet{ribeiro16} pioneered work in this direction with \texttt{LIME}, introducing mappings for images, text, and tabular data. Multiple extensions have subsequently been proposed, recently unified by \citet{lundberg17} with their \texttt{SHAP} tool, in particular its model-agnostic explanation generator \texttt{Kernel SHAP}.

In the time series domain, this explanation approach has not realized its full potential yet. \citet{guilleme2019} gave some first mappings for time series and conducted a study suggesting that the derived explanations are understandable to human users. However, a collection of mappings covering all information available to time series estimators and a thorough discussion of the strengths and weaknesses of the applied methods are still mostly lacking. We introduce the \TXP framework by defining mappings that build on the time domain, the frequency domain, and statistical properties of time series in Section~\ref{sec:framework}. Exemplary explanations generated with \TXP are shown in Figure~\ref{fig:exemplary expl}. We further discuss the assumptions made by our mappings and how they affect the generated explanations in Section~\ref{sec:robust}. Finally, we provide ready-to-use implementations of our framework and existing model-specific approaches as free software and employ them in an experimental comparison in Section~\ref{sec:experiment}.

\section{SHAP Post-hoc Local Explanations}

\begin{figure}[t]
	\centering
	\includegraphics[width=\columnwidth]{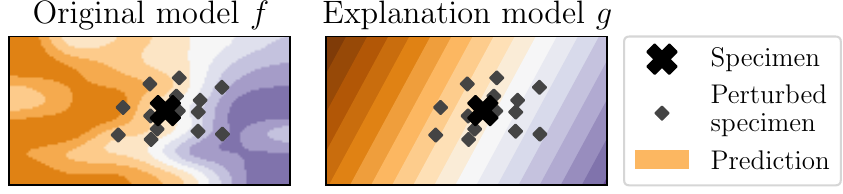}
	\caption{The intuition behind \texttt{SHAP}. The original model $f$ is probed in the vicinity of the specimen. That information is used to build a linear explanation model $g$.}
	\label{fig:local expl intuition}
\end{figure}

\texttt{SHAP} views the black-box ML model to be interpreted as a function $f \colon I \rightarrow \mathbb{R}$ from the input space $I$~\cite{lundberg17}. For a classification task, there is one such function $f^c$ for each class $c$, outputting the probability that the input belongs to $c$. Intuitively, \texttt{SHAP} ``disables'' portions of the specimen $\x \in I$, called \emph{fragments}, yielding a \emph{perturbed specimen} $\z\in I$ and then computes $f(\z)$. Repeating this with different fragments explores the behavior of the model $f$ in the vicinity of the specimen. Using that insight, an interpretable linear model $g$ is built that approximates the original model near $\x$. This is illustrated in Figure~\ref{fig:local expl intuition}. From $g$, we obtain the \emph{impact} of each fragment. It estimates the average $f(\x) - f(\z)$ over all $\z$ with the fragment disabled. We call the estimation quality \emph{fidelity} or \emph{faithfulness} of the impacts to the model $f$. To formalize dividing $\x$ into some $d'$ fragments that can be disabled individually, we introduce the space of \emph{simplified inputs} $I' = \{0, 1\}^{d'}$. Notation with a prime refers to this space. Each 1 or 0 in $\z' \in I'$ respectively stands for an enabled or disabled fragment of the specimen $\x$. The actual disabling of fragments is done by a \emph{mapping function} ${h_{\x} \colon I' \rightarrow I; \z' \mapsto \z}$. We require that evaluating $h_{\x}$ on the all-ones vector $\1 \in I'$ recovers the unperturbed specimen $h_{\x}(\1) = \x$.

\begin{definition}
	\label{def:explanation_model}
	An \emph{explanation model} with respect to a mapping function $h_{\x}$ is a linear regression model $g \colon I' \rightarrow \mathbb{R}$ with the \emph{impact vector} $\p \in \mathbb{R}^{d'}$ such that $g(\1) = f(x)$ and $g(\z') = f(h_{\x}(\0)) + \sum_{k=1}^{d'} \p_k \z'_k$ for all $\z' \in I'$, where $\0 \in I'$ is the zero vector. The set of all possible explanation models is denoted by $G$.
\end{definition}

\noindent
The impact vector $\p$ acts as the local explanation with respect to $\x$. For any fragment index $k \in \{1, \dots, d'\}$, the so-called \emph{SHAP value} $\p_k$ quantifies the impact of the $k$-th fragment. For these impacts to be faithful, $g$ must be \emph{locally faithful} to the model $f$~\cite{ribeiro16}. That is, it approximates $f$ in the vicinity of $\x$, observing $g(\z') \approx f(h_{\x}(\z'))$ if $\z' \in I'$ disables only a few fragments. This is made formal by defining an inverse distance $\pi_{\1} \colon I' \to \mathbb{R}_0^+$. Intuitively, the further away the simplified specimen $\z'$ is from the all-ones vector $\1$, the smaller $\pi_{\1}(\z')$ is. For an appropriate choice of $\pi_{\1}$, we refer the reader to the work of \citet{lundberg17}. The optimal explanation model is then ${g^* = \argmin_{g \in G} \sum_{\z' \in I'} [ f(h_{\x}(\z')) - g(\z') ]^2 \pi_{\1}(\z')}$. By learning the linear regression model $g$ on randomly sampled perturbed simplified specimens $\z'$, \texttt{Kernel SHAP} approximates $g^*$ and obtains the impact vector from $g$.

\section{The timeXplain Mappings}
\label{sec:framework}

While \texttt{Kernel SHAP} can generate model-agnostic explanations, it crucially depends on well-designed mappings for the specific application domain. The fragments of the specimen $\x$ defined by $h_{\x}$ need to be intuitively understandable for the impacts to be meaningful. Mappings not adapted to the data could produce misleading explanations, counteracting the whole idea of interpretability. This makes the construction of good mappings an important challenge.

We now describe the \TXP framework and propose expressive mappings for time series data. It is well-known that time series can be seen both as values over time and as a superposition of frequencies, with the Fourier transform connecting these dual views~\cite{oppenheim10}. Our mappings take both domains as well as the statistics of time series into account.

Throughout this paper, we define the input space $I = \mathbb{R}^d$ to be the set of all time series of length $d$. The value of a series $\z \in I$ at time $t \in \{1, \dots, d\}$ is denoted by $\z_t$.

\subsection{Time Slice Mapping}
\label{sec:time slice mapping}

\begin{figure}[t]
	\centering
	\includegraphics[width=\columnwidth]{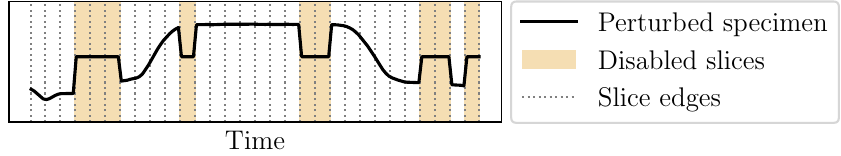}
	\caption{Perturbation of a GunPoint time series using the time slice mapping with global mean replacement.}
	\label{fig:timeslice perturbed}
\end{figure}

We first introduce the \emph{time slice mapping}. The idea is to partition the time series $\x$ into $d'$ contiguous subseries, called \emph{slices}, whose lengths differ by at most $1$. Denote by $\tau \colon \{1, \dots, d\} \to \{1, \dots, d'\}$ the function that assigns to each time $t$ a slice number $\tau(t)$. To disable a slice of $\x$, that is, a fragment of the specimen, one cannot just contract it or fill it with missing values as most models can cope with neither. Instead, a slice is replaced with the corresponding slice of a so-called \emph{replacement time series} $\r \in I$. Five such replacement series will be presented below. An exemplary perturbation (using the global mean replacement) is shown in Figure~\ref{fig:timeslice perturbed}.

\begin{definition}
	Let $\r \in I$ be a \emph{replacement time series}. The \emph{time slice mapping function} $h_{\x,\r} \colon I' \rightarrow I$ yields the perturbed time series that is defined, for all $t \in \{1, \dots, d\}$ and $\z' \in I' = \{0,1\}^{d'}$, as
	\begin{equation*}
		(h_{\x,\r}(\z'))_t =
		\begin{cases}
			\x_t, & \text{if } \z'_{\tau(t)} = 1; \\
			\r_t, & \text{otherwise}.
		\end{cases}
	\end{equation*}
\end{definition}

\noindent
The time slice mapping implicitly assumes that the model bases its decisions solely on the occurrence of patterns in the time domain (\emph{feature space assumption}) and that points in the same slice have similar influence on the model's output (\emph{temporal coherence assumption}). Many, but not all, models from the field of time series estimation fulfill these assumptions. We examine the effects of their violation in Section~\ref{sec:assumptions}.\vspace*{.25em}

\begin{figure}[t]
	\centering
	\includegraphics[width=\columnwidth]{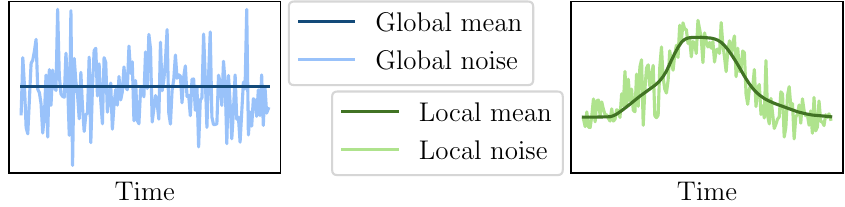}
	\caption{Comparison of four time domain replacement series: global mean/noise and local mean/noise. The UCR GunPoint test set is used as background set.}
	\label{fig:time dom repl}
\end{figure}

\noindent
\textbf{Synthetic Void Information.}
Finding a good replacement series $\r$ is non-trivial. We first propose four \emph{synthetic void information} replacements that carry as little information as possible, see Figure~\ref{fig:time dom repl}. In this setting, replacing a slice of $\x$ with that of $\r$ erases all structure so that the model cannot utilize any information from the ``disabled'' slice. The replacements are defined with respect to an (ideally unbiased) background set $S$ of representative time series, for example test data.

We define the \emph{local mean} replacement by averaging over the $t$-th value in all series in $S$. Averaging over all values in all series gives the \emph{global mean} replacement.
\begin{align*}
	\text{Local mean:} \quad &\r_t^{(1)} = \frac{1}{|S|} \sum_{\s \in S} \s_t;\\
	\text{Global mean:} \quad &\r_t^{(2)} = \frac{1}{|S|} \sum_{\s \in S} \bar{\s}.
\end{align*}

\noindent
For the next two replacement series, we employ Gaussian noise whose parameters are adapted to the series in $S$, again leading to a local and global variant.
\begin{align*}
	\text{Local noise:} \quad &\r_t^{(3)} \sim \mathcal{N}(\mu_t, \sigma_t^2),\
	\mu_t = \r_t^{(1)},\\
	&\sigma_t^2 = \frac{1}{|S|-1} \sum_{\s \in S} (\s_t - \mu_t)^2;\\
	\text{Global noise:} \quad &\r_t^{(4)} \sim \mathcal{N}(\tilde{\mu}, \tilde{\sigma}^2),\
	\tilde{\mu} = \r_0^{(2)},\\
	&\tilde{\sigma}^2 = \frac{1}{|S|(d-1)} \sum_{\s \in S} \sum_{i=1}^d (\s_i - \bar{\s})^2.
\end{align*}

\noindent
Naturally, no replacement may ever be fully void of information. Hence, all of the above replacements are in violation of the \emph{void information assumption} to varying degree. The effects of this are discussed in Section~\ref{sec:assumptions}.\vspace*{.25em}

\noindent
\textbf{Authentic Opposing Information.}
As synthetic void information is never fully void, the model may inadvertently understand it as evidence towards a specific prediction, making explanations hard to interpret. To tackle this issue, we propose \emph{authentic opposing information} replacements. Instead of avoiding any structure, we intentionally insert information different from that of the specimen. We accomplish this by drawing the replacement series $\r^{(5)}$ uniformly from the background set $S$. This has previously been suggested by \citet{guilleme2019}. The impact vector $\p$ now shows how the model discriminates between $\x$ and the particular choice for $\r^{(5)}$. This is easy to interpret, but does not necessarily include all slices responsible for the model's prediction of $\x$. We present a robust modification to fix the latter in Section~\ref{sec:averaging}. Observe that authentic opposing information implicitly assumes \emph{background set homogeneity}, alleging that all series in $S$ are somewhat similar, ensuring that multiple runs yield comparable impacts. For more on this assumption see Section~\ref{sec:assumptions}.

\begin{figure}[t]
	\centering
	\includegraphics[width=\columnwidth]{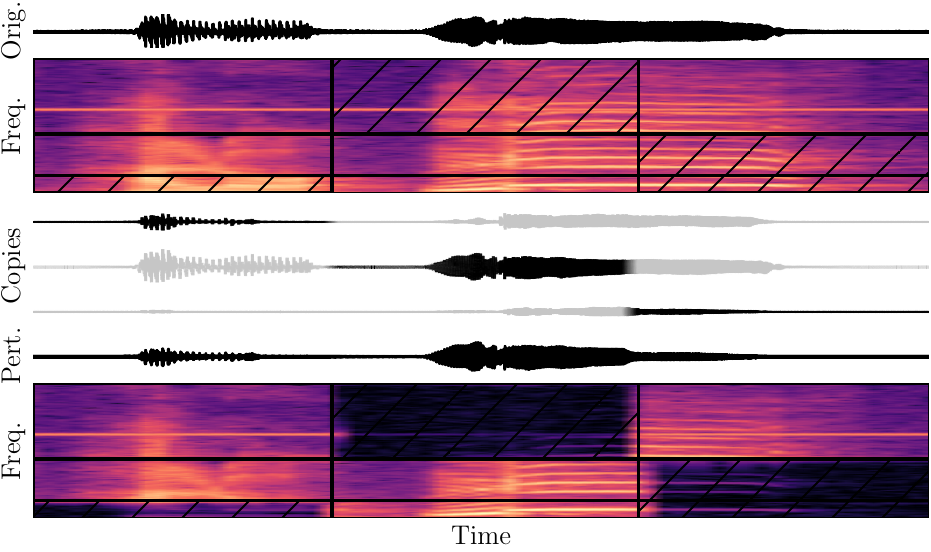}
	\caption{Illustration of the perturbation by the frequency band mapping. The specimen, an audio signal, is depicted in the time domain (``Orig.'') and the frequency domain as a spectrogram (``Freq.''). The spectrogram is diced into nine fragments ($d'_T \,{=}\, 3$ time slices, $d'_B \,{=}\, 3$ frequency bands). For each time slice, a copy of the specimen is created with the respective frequency bands filtered out (Band~1 in the first slice, Band~3 in the second, and Bands~1 \& 2 in the last). The copies are spliced back together, indicated by the shading. The resulting series (``Pert.'') has the hatched fragments blacked out.}
	\label{fig:freqdice perturbed}
\end{figure}

\subsection{Frequency Band Mapping}

Some ML models not only consider the values of a time series but also its frequency spectrum. To explain those decisions as well, we need to introduce a mapping that considers the frequency domain. The \emph{frequency band mapping} is motivated by spectrograms, which reveal the magnitudes of the frequencies that make up a signal around each time stamp. It is illustrated in Figure~\ref{fig:freqdice perturbed}.

We split the time series $\x$ into $d'_T$ time slices, whose spectra are respectively split into $d'_B$ frequency bands with quadratically increasing bandwidths, yielding a total of $d' = d'_T \,{\cdot}\, d'_B$ fragments. A quadratic scaling gives finer resolution at the information-rich lower frequencies and is not too coarse at higher frequencies, as opposed to a log scaling. The mapping receives a simplified input matrix $\z' \in I' = \{0, 1\}^{d'_B \times d'_T}$. The $\tau$-th column $\z'_{\bullet,\tau}$ indicates which frequency bands to disable in the $\tau$-th time slice, we call this a \emph{band profile}. A \emph{strategy function} $\sigma_{\x} \colon \{0, 1\}^{d'_B} \rightarrow I$ then specifies how the bands are disabled in $\x$. We propose three strategies below. Function $\sigma_{\x}$ is queried once for each profile, yielding $d'_T$ perturbed copies of $\x$. The copies are spliced together such that the $\tau$-th copy is active in the $\tau$-th slice. We use linear cross-fades, which do not cause a dip in signal power here since the perturbed copies are typically still highly correlated \cite{fink16}.

This mapping relaxes two assumptions made by the time slice mapping. Firstly, the model may now use both temporal and spectral information (weakened feature space assumption), and secondly, neighboring temporal features, for example waves, must now only have similar influence on the predictions of the model when they lie in the same spectral region (weakened coherence assumption). See Section~\ref{sec:assumptions} for details.\vspace*{.25em}

\noindent
\textbf{Synthetic Void Information.}
Again, we first propose a synthetic void information approach. When queried with some band profile $\z'_{\bullet,\tau} \in \{0, 1\}^{d'_B}$, the \emph{filter} strategy function $\sigma_{\x}$ aims to zero the magnitudes of those frequency bands $\beta$ of the specimen $\x$ for which $\z'_{\beta,\tau} = 0$ using so-called \emph{bandstop filters}. In Figure~\ref{fig:freqdice perturbed}, the disabled fragments are blacked out in the spectrogram while others stay intact. Filters need to find a trade-off between dampening only the magnitudes of the desired frequency bands and inserting undesired artifacts. We compare two types of filters from digital signal processing. An \emph{elliptic filter} is an infinite impulse response (IIR) filter with sharp cutoffs and small computational complexity. However, IIR filters are notorious for their unforeseen artifacts and require manual tests on the specimens in advance~\cite{oppenheim10}. We recommend using a finite impulse response filter with least-squares optimization (FIRLS). While it may not attain the same cutoff sharpness as an elliptic filter, its reliability makes it a robust choice for the automated generation of explanations for unknown specimens.\vspace*{.25em}

\noindent
\textbf{Authentic Opposing Information.}
As models might be misled by zeroed frequency magnitudes, we also propose an authentic opposing information approach. The \emph{patch strategy} alters the magnitudes and phases inside disabled bands to those of a uniformly drawn \emph{patch time series} $\bm{r} \in S$. It assumes background set homogeneity.

Let $\RDFT \colon I \rightarrow \mathbb{C}^{1 + \lfloor\sfrac{d}{2}\rfloor}$ denote the discrete Fourier transform on real inputs~\cite{Sorensen87realDFT}. In the resulting series, the fluctuations of values are associated with the change of frequency. Coupled with the inverse transform ${\iRDFT \colon \mathbb{C}^{1 + \lfloor\sfrac{d}{2}\rfloor} \rightarrow I}$, we can manipulate time series in the frequency domain. Let the function $\beta \colon \{ 1, \dots, \lfloor\sfrac{d}{2}\rfloor \} \rightarrow \{1, \dots, d'_B\}$ assign to each frequency bin $\omega \neq 0$ the number of its enclosing band.

\begin{definition}
	Let $\r \in S$ be a \emph{patch time series}. The \emph{patch strategy function} $\sigma_{\x,\r} \colon \{0,1\}^{d'_B} \rightarrow I$ yields the perturbed time series which is defined, for all band profiles $\z'_{\bullet,\tau} \in \{0,1\}^{d'_B}$ and $\omega \in \{0, \dots, \lfloor\sfrac{d}{2}\rfloor\}$, as
	\begin{align*}
		\sigma_{\x,\r}(\z'_{\bullet,\tau}) &= \iRDFT(\bm{Z}) \quad\ \,\, \text{with}\\
		\bm{Z}_{\omega} &=
		\begin{cases}
			(\RDFT(\x))_{\omega}, & \text{if } \omega = 0 \text{ or } \z'_{\beta(\omega),\tau} = 1; \\
			(\RDFT(\r))_{\omega}, & \text{otherwise}.
		\end{cases}
	\end{align*}
\end{definition}

\subsection{Statistics Mapping}

Some ML models may consider the mean and variance of a time series. To detect this, we propose the \emph{statistics mapping} $h_{\x,\mu,\sigma^2}$ for replacement values $\mu$, $\sigma^2$. Simplified inputs $\z'$ now have only $d' = 2$ positions. If $\z'_1 = 0$, then $h_{\x,\mu,\sigma^2}(\z')$ is a re-normalization of the specimen $\x$ such that its mean equals the replacement mean $\mu$; same for the variance $\sigma^2$ if $\z'_2 = 0$. The replacements are devised from a background set $S$ as in Section~\ref{sec:time slice mapping} (synthetic void information), or by using a uniformly drawn time series from $S$ (authentic opposing information).

\section{Robustness of timeXplain}
\label{sec:robust}

Employing the above mappings without considering the application context may result in misleading explanations. We highlight the most common pitfalls and how to avoid them, in particular regarding the assumptions.

\subsection{Multi-Run Averaging and Background Set Class Splitting}
\label{sec:averaging}

Impact vectors from authentic opposing information reveal how the model discerns $\x$ from one specific series $\r \in S$. To find more fragments that are influential in the prediction of $\x$ and to overcome inhomogeneities in $S$, we can instead ask the model to discern $\x$ from $n$ different picks from $S$ and then average the impacts.

For classification, drawing uniformly or devising a replacement, such as local mean, from the background set $S$ favors classes dominant in $S$. Also, a replacement devised from series of different classes might not be meaningful. We combat both issues by treating the classes separately. A classifier has a probability model $f^c$ for each class. To derive the impacts $\bm{\phi}^c$ for $f^c$, we average intermediate impacts $\bm{\phi}^{c,\tilde{c}}$ for \emph{background classes} $\tilde{c}$.

Both for classification and other TSE tasks, we express the generation of the impact vectors (Definition~\ref{def:explanation_model}) as an operator $\Shap(f, h_{\x,\R})$. The replacement information $\R$ is devised either from a whole background set $T$ or from a single series $\r$, denoted by $\R(T)$ or $\R(\r)$, respectively. For classification, let $C$ be the set of all classes, and let $S^c \subseteq S$ contain the series with true class $c$.

\begin{definition}
	Let $f$ be a non-classifier. If $\R$ is to be devised from a set, the \emph{impact vector} is $\bm{\phi} = \Shap(f, h_{\x,\R(S)})$. If instead $\R$ is to be devised from a single series, let $\r^1, \r^2, \dots, \r^n$ be distinct uniform samples from $S$. The \emph{impact vector} is the average $\bm{\phi} = \tfrac{1}{n} \sum_{i=1}^n \Shap(f, h_{\x,\R(\r^i)})$.

	For classifiers, let $c \in C$ be a class and $f^c$ the corresponding model. We define the \emph{intermediate impact vector} $\p^{c,\tilde{c}}$ for the \emph{background class} $\tilde{c} \in C$ as follows. If $\R$ is to be devised from a set, then $\p^{c,\tilde{c}} = \Shap(f^c, h_{\x,\R(S^{\tilde{c}})})$. If instead $\R$ is to be devised from a single series, let $\r^1, \r^2, \dots, \r^n$ be distinct uniform samples from $S^{\tilde{c}}$ and $\p^{c,\tilde{c}} = \tfrac{1}{n} \sum_{i=1}^n \Shap(f^c, h_{\x,\R(\r^i)})$. The final \emph{impact vector} for class $c$ is $\p^c = \tfrac{1}{|C|} \sum_{\tilde{c} \in C} \p^{c,\tilde{c}}$.
\end{definition}

\subsection{Artifacts from False Assumptions}
\label{sec:assumptions}

All mappings make various assumptions. We explore what happens if they are violated in order to help the user assess the merits of \TXP explanations.\vspace*{.25em}

\noindent
\textbf{Feature Space Assumption.}
This assumption alleges that the model bases its decisions on a specific subset of temporal, spectral, or statistical properties of the specimen. What happens if it is violated, for example by using the time slice mapping with a model that only considers the spectrum of the specimen? When disabling a time slice with a spike, the magnitude of the corresponding frequency reduces. If this frequency is influential to the model, its prediction changes. Hence, the time slice gains impact even though it is not directly influential. The end user interpreting the impact vector needs to be aware that the high-impact fragments might not be the influential properties, but instead only artifacts. Conversely, the user may need additional knowledge of the model to choose an appropriate mapping.\vspace*{.25em}

\begin{figure}[t]
	\centering
	\includegraphics[width=\columnwidth]{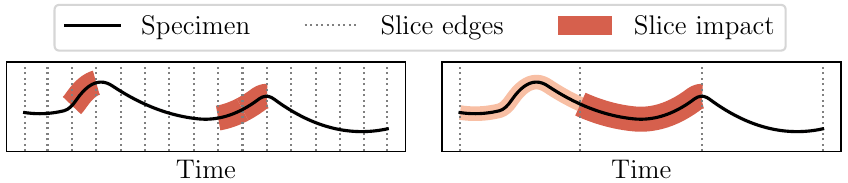}
	\caption{The left plot assumes only short-reaching coherence by using many slices. The two rising slopes have a high impact. The right plot assumes far-reaching coherence, resulting in a misleading impact distribution.}
	\label{fig:coherence assumption}
\end{figure}

\noindent
\textbf{Coherence Assumption.}
Assuming coherence alleges that close-by time stamps and/or frequencies have similar influence on the model's prediction. Intuitively, this is reasonable as a model that bases its decision on isolated time stamps or frequencies would be overfitting.

Contrarily, far-reaching coherence becomes increasingly implausible and assuming it yields misleading explanations if influential portions of the specimen share fragments with non-influential portions, as illustrated in Figure~\ref{fig:coherence assumption}. The user must be aware that the impact of a fragment only reveals the approximate sum of the impacts of any possible subdivision of the fragment.\vspace*{.25em}

\noindent
\textbf{Simplified Input Dimension Trade-off.}
Above, we saw that a too small number of fragments $d'$ results in low resolution and misleading impacts. Intuitively, the larger we choose $d'$, the more fine-grained the explanation becomes. However, it quickly becomes prohibitive to query the model with all $2^{d'}$ perturbations of a specimen to generate just a single explanation. \texttt{Kernel SHAP} thus resorts to sampling. For large $d'$, too few perturbations would be sampled and the explanation model $g$ would overfit, rendering the impacts inaccurate.

In summary, $d'$ being either too low or too high may both reduce the quality of the explanation. At this time, the best way of choosing a good $d'$ may be to manually select the highest $d'$ whose impacts are sufficiently stable over multiple tries. In this work, we choose $d'$ based on such preliminary experiments. A thorough investigation of $d'$ is an interesting avenue for future work.\vspace*{.25em}

\noindent
\textbf{Void Information Assumption.}
The impact of a fragment estimates the average decrease of the model's output when disabling that fragment. If we could use perfectly void information when disabling, the impacts would be easy to interpret. However, no synthetic void information is perfect and the model may inadvertently understand it as evidence towards a specific prediction. The more the influence of this evidence on the model is unknown, the more it becomes unclear how to interpret the impacts. In contrast, when using authentic opposing information, the influence is known exactly, and hence, such impacts are in general more interpretable.

\noindent
\textbf{Background Set Homogeneity.}
The background set $S$ is likely to be heterogeneous. Even in the same class it may range from shifted series to completely distinct ones. When using authentic opposing replacement information, that is, uniform samples from $S$, this reduces the robustness of the impact computation over multiple tries. It may even neutralize multi-run averaging, see Section~\ref{sec:averaging}. When the background set is very heterogeneous, synthetic void information should be preferred.

\section{Experimental Evaluation}
\label{sec:experiment}

In this experimental section, we assess \TXP with the time slice mapping when explaining the predictions of state-of-the-art time series classifier. We aim to identify the best replacements and whether to employ background set class splitting. We also compare our results to model-specific explanation methods. The experiment is conducted by first generating impacts using all explanation methods and then evaluating their quality using two metrics. Due to space constraints, we leave the evaluation of the frequency band mapping for future work.

\subsection{Classifiers}

We include classifiers employing different paradigms, as outlined by \citet{bagnall17}. \emph{Interval techniques} analyze training-selected, phase-independent intervals of a series, ignoring noisy regions and focusing on discriminatory ones. We choose \emph{time series forests} (\texttt{TSF}) as representative. \emph{Shapelet techniques}, like \emph{shapelet transform} with a random forest (\texttt{ST}), classify series based on the occurrence of class-specific, phase-independent subseries found during training, called \emph{shapelets}. Finally, \emph{dictionary-based techniques} approximate subseries as words that then form n-grams, whose counts are fed into a classifier. From the time domain, we choose the \emph{symbolic aggregate approximation vector space model} (\texttt{SAX-VSM}) of \citet{senin13}. From the frequency domain, we choose \emph{word extraction for time series classification} with a logistic regressor (\texttt{WEASEL}) by \citet{schafer17}. On top, we include three general purpose classifiers which are known to perform well on time series data~\cite{bagnall17}, namely, the \emph{rotation forest} (\texttt{RotF}), a \emph{support vector machine} with a linear kernel (\texttt{SVM/Lin}), and a \emph{residual network} (\texttt{ResNet}) as the current best representative of deep learning on time series according to~\citet{fawaz19}.

Some of these allow for model-specific explanation methods that assign an impact to each time stamp of the time series. Predictions of linear and decision tree models can be explained with \texttt{Linear SHAP}~\cite{lundberg17} and \texttt{Tree SHAP}~\cite{lundberg20}, respectively. For neural networks with a global average pooling (GAP) layer, class activation maps (CAM) have been used~\cite{fawaz19}. They omit the averaging, recovering the contribution of each time stamp to each output neuron. Shapelet classifiers are explainable by first finding the impact of each shapelet, for example via \texttt{Tree SHAP}, and then adding (``superpositioning'') the impact of any shapelet in the interval of its occurrence~\cite{hills13shapeletJournal}. Symbolic aggregate approximation and symbolic Fourier approximation models allow for finding the impact of each n-gram and then ``superpositioning'' their occurrences~\cite{nguyen18}.

\subsection{Evaluation Metrics}

Recall that \TXP assigns impacts $\p \in \mathbb{R}^{d'}$ to time slices (\emph{slice-wise}) while model-specific explanations assign impacts $\bm{\psi} \in I \,{=}\, \mathbb{R}^d$ to time stamps (\emph{stamp-wise}). One can convert slice-wise to stamp-wise impacts by setting the impact of a time stamp to the impact of its surrounding slice divided by the slice length. We evaluate stamp-wise explanations using two measures. We assume $f$ is the model, $\x \in I$ is the specimen, and $\bm{\psi} \in I$ is a stamp-wise impact vector that explains $f(\x)$.\vspace*{.25em}

\noindent
\textbf{Dynamic Time Warping Interval Fidelity.}
We aim to assess the fidelity of stamp-wise impacts. To not disadvantage model-specific explanations, we relax it in a way that when ``disabling'' time stamps, the model's output is only required to decrease \emph{proportionally} to the sum of impacts of the stamps. To quantify this, we first partition the interval $[1,d]$ into $k-1$ subintervals of integral size whose size difference is at most one. Then, at each of the $k$ endpoints, we compute an evaluation function $\alpha$ and integrate it using the trapezoidal rule.

Given some endpoint $\ell$, to compute $\alpha(\ell)$ consider $m$ evenly spread, potentially overlapping intervals $[a_{i}, b_{i}]$, $i \in \{1, \dots, m\}$, $a_{i}, b_{i} \in \{1, \dots, d\}$ of length $\ell$ over $\x$. Imagine that we have, for each interval $[a_i,b_i]$, some time series $\w^i$ that matches $\x$ outside of the interval and is as different from $\x$ as possible inside. Then, $\alpha(\ell)$ is defined as the Pearson correlation of the sequences $(\sum_{t=a_{i}}^{b_{i}} \bm{\psi}_t )_{i=1}^m$ and $(f(\x) - f(\bm{w}^{i} ))_{i=1}^m$.

We are left to describe how to obtain $\w^i$. Let $t_{\x}$ and $t_{\z}$ denote time stamps of the respective series $\x$ and $\z$. We first apply to all series $\z \in S$ a modified \emph{dynamic time warping} (DTW) algorithm given by
\begin{align*}
	\text{DTW}_{i,\z}(t_{\x}, t_{\z}) &= \min \{ \text{DTW}_{i,\z}(t_{\x}, t_{\z} \,{-}\, 1),\\
	&\hspace*{0.275em}\phantom{= \min \{ }\text{DTW}_{i,\z}(t_{\x} \,{-}\, 1, t_{\z}) + \delta,\\
	&\hspace*{0.275em}\phantom{= \min \{ }\text{DTW}_{i,\z}(t_{\x} \,{-}\, 1, t_{\z} \,{-}\, 1) + \delta \};\\[.25em]
	\text{DTW}_{i,\z}(-1, t_{\z}) &= 0;\ \text{DTW}_{i,\z}(t_{\x},-1) = \infty;\\
	\delta &=
	\begin{cases}
		\left| \x_{t_{\x}} \,{-}\, \z_{t_{\z}} \right|, &\text{if } t_{\x} \notin [a_{i}, b_{i}]; \\
		-\sqrt{\left| \x_{t_{\x}} \,{-}\, \z_{t_{\z}} \right|}, &\text{if } t_{\x} \in [a_{i}, b_{i}].
	\end{cases}	
\end{align*}

\noindent
We preempt heavy warping with a \emph{Sakoe-Chiba band} of width $\lambda$~\cite{sakoe78}. To obtain $\w^i$, we pick the series $\z$ minimizing $\text{DTW}_{i,\z}(d, d)$ and recover its warped version $\w^{i}$ by backtracking through the DTW matrix, storing $\z_{t_{\z}}$ whenever $t_{\x}$ is about to change.

Preliminary experiments have shown that choosing $\lambda$ as $2 \lfloor \sfrac{d}{10} \rfloor + 1$ yields suitable series $\w^{i}$. For the resolution, we choose $k = 10$ endpoints and $m = 50$ intervals.

Notice that our metric does not synthetically perturb $\x$ in the intervals, but instead finds ``real'' time series that represent the perturbation, this avoids bias towards a specific replacement. The noise introduced by this extra step is negligible as we average over sufficiently many specimens and, for a fixed specimen, use the same set of $\w^{i}$ for each evaluated impact vector.\vspace*{.25em}

\noindent
\textbf{Informativeness.}
\citet{nguyen20} introduced the \emph{informativeness} quality measure in the context of classification. We adapt their idea. Let $\tilde{\sigma}^2$ be the variance of a background set $S$ as in the global noise replacement (Section~\ref{sec:time slice mapping}). For each $q \in \{0, 1, \dots, 100\}$, find the time stamps with the $q$\% highest impacts in $\bm{\psi}$ and perturb $\x$ by adding Gaussian noise with variance $\tilde{\sigma}^2$ to those time stamps, yielding $\z_q \in I$. Integrating $q \mapsto f(\z_q)$ using the trapezoidal rule gives the score.

\subsection{Experimental Setup}

\begin{figure}[t]
	\centering
	\includegraphics[width=\columnwidth]{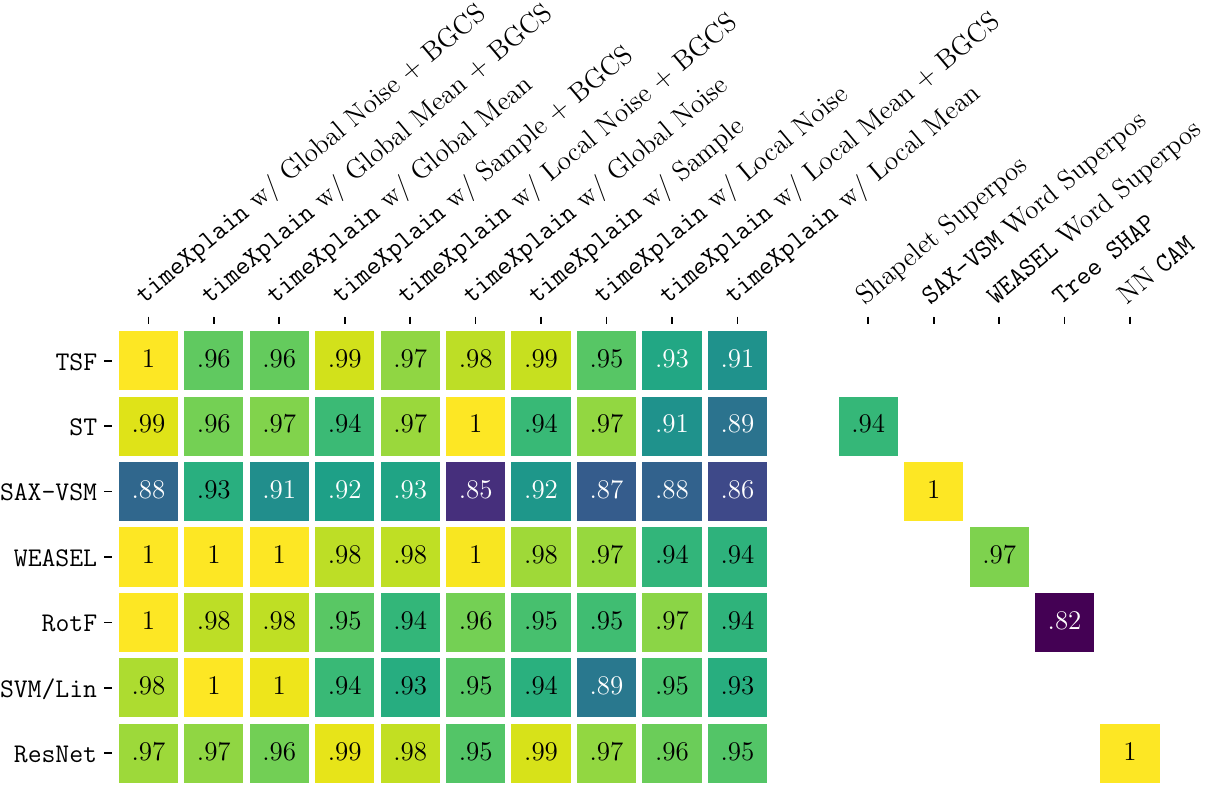}
	\caption{Mean DTW interval fidelity for classifiers and explanation methods. Each row is normalized to max~1. ``BGCS'' stands for ``background set class splitting''.}
	\label{fig:fidelity matrix}
\end{figure}

The classifiers were implemented in Python using the packages \texttt{scikit-learn}~v0.21.2 \cite{scikit-learn}, \texttt{sktime}~v0.3.0 \cite{loning19}, \texttt{pyts}~v0.8.0 \cite{faouzi20}, \texttt{tslearn}~v0.1.29 \cite{tavenard17}, and \texttt{tensorflow}~v1.14.0 \cite{tensorflow}. We used a server with four Intel\textsuperscript{\textregistered} Xeon\textsuperscript{\textregistered} Gold 5118 CPUs at 2.3~GHz, 62~GB RAM, and the Ubuntu v18.04.3 OS. The data was taken from the UCR time series classification archive~\cite{ucr18}. We excluded datasets with missing values or varying series lengths as some models cannot cope with them. We then selected 61 datasets at random and trained all models on each one.

From each dataset's test set, five specimens were selected at random in a way that specimen~1 is of class~1, specimen~2 of class~2, and so on. For each specimen $\x$ with true class $c_{\x}$ and each classifier given by functions $f^c$ for class $c$, impact vectors that explain $f^{c_{\x}}(\x)$ were computed using the explanation methods mentioned at the beginning of Section~\ref{sec:experiment}. For \TXP, the test sets served as background sets $S$, the number of time slices $d'$ was set to $\min \{ \lfloor{\sfrac{d}{5}}\rfloor, 30 \}$, we employed multi-run averaging with $n = 10$ runs, and the models were probed with 1000 perturbations per run. All impact vectors were evaluated with both metrics.

\begin{figure}[t]
	\centering
	\includegraphics[width=\columnwidth]{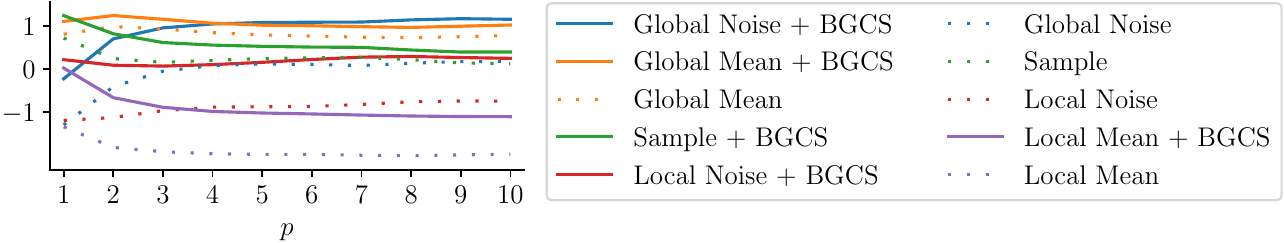}
	\caption{Progression of the mean DTW interval fidelity for all time slice replacements when only integrating up to the $p$-th $\alpha(\ell)$. For each $p$, the values are z-normalized.}
	\label{fig:fidelity progression}
\end{figure}

\subsection{Results}

Figure~\ref{fig:fidelity matrix} shows the mean DTW interval fidelity. Here, \TXP with global mean or noise replacement is best over most classifiers and even outperforms some  model-specific explanation methods. Local noise and especially local mean are worst as they keep a lot of structure intact. For \texttt{ResNet}, only sample replacement is as good as the model-specific \texttt{CAM}, and only for \texttt{SAX-VSM} \TXP is substantially worse than a model-specific approach. Employing background set class splitting almost always improves the fidelity. The informativeness is not shown as it did not differ much over the methods.

Figure~\ref{fig:fidelity progression} displays how the mean fidelity develops as we integrate up to increasing interval lengths, irrespective of the classifier. Here, background set class splitting is always an improvement. For small lengths, global noise lacks fidelity; same for sample with large lengths.

As sample replacement is easiest to interpret and has suitable fidelity, we recommend defaulting to it. Global noise is only better when extensive fidelity is important. One should always use background set class splitting.

\section{Conclusion}

The ubiquity of machine learning systems in our everyday life raises the need to interpret them. We devised \TXP, a model-agnostic post-hoc local explanation framework for time series estimators. We introduced several domain mappings building on the \texttt{SHAP} framework. These mappings were embedded in a system of categories, depending on whether they use time slices, frequency bands, or time series statistics, and whether they employ synthetic void information or authentic opposing one. This categorization makes it easy to extend and adapt the framework. We also discussed the implicit assumptions underlying the mappings and how they affect the robustness of \TXP. We have experimentally shown that our time slice mapping with a sample or global noise replacement produces explanations often more faithful to the model than other replacements and even some model-specific approaches.

There are two main avenues for future work. One is a thorough evaluation of \TXP in the frequency domain similar to our time domain analysis here. This should be accompanied by a novel evaluation metric and novel model-specific spectral explanation methods. Another interesting direction would be to employ \TXP in a large-scale comparison of time series classifiers. This may uncover unexpected similarities between models differing in design and focus.

\section*{Ethics Statement}

The core of the presented work is to enable researchers and practitioners to access the predictions of time series estimators that previously only appeared as black boxes. This allows them to detect any unwanted bias or even malicious discrimination. It is thus a crucial technical prerequisite in order to make artificial intelligence systems adhere to ethical standards and legal regulations. Unfortunately, our proposed method is computationally expensive which comes with a significant environmental footprint, especially in energy consumption.

\section*{Acknowledgements}

We thank Fabian Geier, Emmanuel Müller, and Erik Scharwächter for first posing the research question of how LIME, and by extension SHAP, could be applied to time series data. Additionally, we are thankful to Jannes Münchmeyer and Mario Sänger for many interesting discussions, without which this work would not have been possible. Finally, we would like to thank Jorin Heide, Linus Heinzl, Nicolas Klodt, Lars Seifert, and Arthur Zahn for proofreading early drafts of this paper, as well as Sarah Shtaierman for her final proofreading. This work was partially supported by the ILB ProFIT project ``Virtual Compressor'' under agreement no.~80173319.

\bibliography{references}

\begin{thebibliography}{31}
\providecommand{\natexlab}[1]{#1}
\providecommand{\url}[1]{\texttt{#1}}
\providecommand{\urlprefix}{URL }
\expandafter\ifx\csname urlstyle\endcsname\relax
  \providecommand{\doi}[1]{doi:\discretionary{}{}{}#1}\else
  \providecommand{\doi}{doi:\discretionary{}{}{}\begingroup
  \urlstyle{rm}\Url}\fi

\bibitem[{Abadi et~al.(2015)Abadi, Agarwal, Barham, Brevdo, Chen, Citro,
  Corrado, Davis, Dean, Devin, Ghemawat, Goodfellow, Harp, Irving, Isard, Jia,
  Jozefowicz, Kaiser, Kudlur, Levenberg, Man\'{e}, Monga, Moore, Murray, Olah,
  Schuster, Shlens, Steiner, Sutskever, Talwar, Tucker, Vanhoucke, Vasudevan,
  Vi\'{e}gas, Vinyals, Warden, Wattenberg, Wicke, Yu, and Zheng}]{tensorflow}
Abadi, M.; Agarwal, A.; Barham, P.; Brevdo, E.; Chen, Z.; Citro, C.; Corrado,
  G.~S.; Davis, A.; Dean, J.; Devin, M.; Ghemawat, S.; Goodfellow, I.; Harp,
  A.; Irving, G.; Isard, M.; Jia, Y.; Jozefowicz, R.; Kaiser, L.; Kudlur, M.;
  Levenberg, J.; Man\'{e}, D.; Monga, R.; Moore, S.; Murray, D.; Olah, C.;
  Schuster, M.; Shlens, J.; Steiner, B.; Sutskever, I.; Talwar, K.; Tucker, P.;
  Vanhoucke, V.; Vasudevan, V.; Vi\'{e}gas, F.; Vinyals, O.; Warden, P.;
  Wattenberg, M.; Wicke, M.; Yu, Y.; and Zheng, X. 2015.
\newblock TensorFlow: Large-Scale Machine Learning on Heterogeneous Systems.

\bibitem[{Alvarez-Melis and Jaakkola(2018)}]{alvarez18selfexpl}
Alvarez-Melis, D.; and Jaakkola, T.~S. 2018.
\newblock Towards Robust Interpretability With Self-Explaining Neural Networks.
\newblock In \emph{Advances in Neural Information Processing Systems, NeurIPS}.

\bibitem[{Bagnall et~al.(2017)Bagnall, Lines, Bostrom, Large, and
  Keogh}]{bagnall17}
Bagnall, A.; Lines, J.; Bostrom, A.; Large, J.; and Keogh, E. 2017.
\newblock The Great Time Series Classification Bake Off: A Review and
  Experimental Evaluation of Recent Algorithmic Advances.
\newblock \emph{Data Mining and Knowledge Discovery} 31(3).

\bibitem[{Dau et~al.(2018)Dau, Keogh, Kamgar, Yeh, Zhu, Gharghabi,
  Ratanamahatana, Yanping, Hu, Begum, Bagnall, Mueen, and Batista}]{ucr18}
Dau, H.~A.; Keogh, E.; Kamgar, K.; Yeh, C.-C.~M.; Zhu, Y.; Gharghabi, S.;
  Ratanamahatana, C.~A.; Yanping; Hu, B.; Begum, N.; Bagnall, A.; Mueen, A.;
  and Batista, G. 2018.
\newblock The UCR Time Series Classification Archive.

\bibitem[{Du, Liu, and Hu(2020)}]{Du20Communications}
Du, M.; Liu, N.; and Hu, X. 2020.
\newblock Techniques for Interpretable Machine Learning.
\newblock \emph{Communications of the ACM} 63(1).

\bibitem[{Faouzi and Janati(2020)}]{faouzi20}
Faouzi, J.; and Janati, H. 2020.
\newblock pyts: A Python Package for Time Series Classification.
\newblock \emph{Journal of Machine Learning Research} 21(46).

\bibitem[{Fawaz et~al.(2019)Fawaz, Forestier, Weber, Idoumghar, and
  Muller}]{fawaz19}
Fawaz, H.~I.; Forestier, G.; Weber, J.; Idoumghar, L.; and Muller, P.-A. 2019.
\newblock Deep Learning for Time Series Classification: A Review.
\newblock \emph{Data Mining and Knowledge Discovery} 33(4).

\bibitem[{Fink, Holters, and Z{\"o}lzer(2016)}]{fink16}
Fink, M.; Holters, M.; and Z{\"o}lzer, U. 2016.
\newblock Signal-Matched Power-Complementary Cross-Fading and Dry-Wet Mixing.
\newblock In \emph{International Conference on Digital Audio Effects, DAFx}.

\bibitem[{Goodman and Flaxman(2017)}]{goodman17}
Goodman, B.; and Flaxman, S. 2017.
\newblock European Union Regulations on Algorithmic Decision Making and a
  ``Right to Explanation''.
\newblock \emph{AI Magazine} 38(3).

\bibitem[{Guillem{\'e} et~al.(2019)Guillem{\'e}, Masson, Roz{\'e}, and
  Termier}]{guilleme2019}
Guillem{\'e}, M.; Masson, V.; Roz{\'e}, L.; and Termier, A. 2019.
\newblock Agnostic Local Explanation for Time Series Classification.
\newblock In \emph{IEEE International Conference on Tools with Artificial
  Intelligence, ICTAI}.

\bibitem[{Hills et~al.(2014)Hills, Lines, Baranauskas, Mapp, and
  Bagnall}]{hills13shapeletJournal}
Hills, J.; Lines, J.; Baranauskas, E.; Mapp, J.; and Bagnall, A. 2014.
\newblock Classification of Time Series by Shapelet Transformation.
\newblock \emph{Data Mining and Knowledge Discovery} 28(4).

\bibitem[{Lin et~al.(2019)Lin, Hsieh, Cheng, Huang, and
  Adnan}]{Lin19predictiveMaintenance}
Lin, C.; Hsieh, Y.; Cheng, F.; Huang, H.; and Adnan, M. 2019.
\newblock Time Series Prediction Algorithm for Intelligent Predictive
  Maintenance.
\newblock \emph{IEEE Robotics and Automation Letters} 4(3).

\bibitem[{Lipton(2018)}]{lipton16}
Lipton, Z.~C. 2018.
\newblock The Mythos of Model Interpretability.
\newblock \emph{ACM Queue} 16(3).

\bibitem[{L{\"{o}}ning et~al.(2019)L{\"{o}}ning, Bagnall, Ganesh, Kazakov,
  Lines, and Kir{\'{a}}ly}]{loning19}
L{\"{o}}ning, M.; Bagnall, A.; Ganesh, S.; Kazakov, V.; Lines, J.; and
  Kir{\'{a}}ly, F.~J. 2019.
\newblock sktime: A Unified Interface for Machine Learning with Time Series.
\newblock In \emph{Workshop on Systems for ML, MLSys}.

\bibitem[{Lou, Caruana, and Gehrke(2012)}]{lou12}
Lou, Y.; Caruana, R.; and Gehrke, J. 2012.
\newblock Intelligible Models for Classification and Regression.
\newblock In \emph{ACM International Conference on Knowledge Discovery and Data
  Mining, SIGKDD}.

\bibitem[{Lundberg et~al.(2020)Lundberg, Erion, Chen, DeGrave, Prutkin, Nair,
  Katz, Himmelfarb, Bansal, and Lee}]{lundberg20}
Lundberg, S.~M.; Erion, G.; Chen, H.; DeGrave, A.; Prutkin, J.~M.; Nair, B.;
  Katz, R.; Himmelfarb, J.; Bansal, N.; and Lee, S.-I. 2020.
\newblock From Local Explanations to Global Understanding with Explainable AI
  for Trees.
\newblock \emph{Nature Machine Intelligence} 2(1).

\bibitem[{Lundberg and Lee(2017)}]{lundberg17}
Lundberg, S.~M.; and Lee, S.~I. 2017.
\newblock A Unified Approach to Interpreting Model Predictions.
\newblock In \emph{Advances in Neural Information Processing Systems, NeurIPS}.

\bibitem[{Nguyen et~al.(2018)Nguyen, Gsponer, Ilie, and Ifrim}]{nguyen18}
Nguyen, T.~L.; Gsponer, S.; Ilie, I.; and Ifrim, G. 2018.
\newblock Interpretable Time Series Classification using All-Subsequence
  Learning and Symbolic Representations in Time and Frequency Domains.
\newblock \emph{arXiv preprint arXiv:1808.04022} .

\bibitem[{Nguyen et~al.(2019)Nguyen, Gsponer, Ilie, O'Reilly, and
  Ifrim}]{nguyen19InterpretableLinearJournal}
Nguyen, T.~L.; Gsponer, S.; Ilie, I.; O'Reilly, M.; and Ifrim, G. 2019.
\newblock Interpretable Time Series Classification using Linear Models and
  Multi-resolution Multi-domain Symbolic Representations.
\newblock \emph{Data Mining and Knowledge Discovery} 33(4).

\bibitem[{Nguyen, Nguyen, and Ifrim(2020)}]{nguyen20}
Nguyen, T.~T.; Nguyen, T.~L.; and Ifrim, G. 2020.
\newblock A Model-Agnostic Approach to Quantifying the Informativeness of
  Explanation Methods for Time Series Classification.
\newblock In \emph{Workshop on Advanced Analytics and Learning on Temporal
  Data, AALTD}.

\bibitem[{Oppenheim and Schafer(2010)}]{oppenheim10}
Oppenheim, A.~V.; and Schafer, R.~W. 2010.
\newblock \emph{Discrete-Time Signal Processing}.
\newblock Prentice-Hall Signal Processing Series. Pearson.

\bibitem[{Pedregosa et~al.(2011)Pedregosa, Varoquaux, Gramfort, Michel,
  Thirion, Grisel, Blondel, Prettenhofer, Weiss, Dubourg, Vanderplas, Passos,
  Cournapeau, Brucher, Perrot, and Duchesnay}]{scikit-learn}
Pedregosa, F.; Varoquaux, G.; Gramfort, A.; Michel, V.; Thirion, B.; Grisel,
  O.; Blondel, M.; Prettenhofer, P.; Weiss, R.; Dubourg, V.; Vanderplas, J.;
  Passos, A.; Cournapeau, D.; Brucher, M.; Perrot, M.; and Duchesnay, E. 2011.
\newblock scikit-learn: Machine Learning in Python.
\newblock \emph{Journal of Machine Learning Research} 12.

\bibitem[{Rajkomar et~al.(2018)Rajkomar, Oren, Chen, Dai, Hajaj, Liu, Liu, Sun,
  Sundberg, Yee, Zhang, Duggan, Flores, Hardt, Irvine, Le, Litsch, Marcus,
  Mossin, and Dean}]{Rajkomar18health}
Rajkomar, A.; Oren, E.; Chen, K.; Dai, A.; Hajaj, N.; Liu, P.; Liu, X.; Sun,
  M.; Sundberg, P.; Yee, H.; Zhang, K.; Duggan, G.; Flores, G.; Hardt, M.;
  Irvine, J.; Le, Q.; Litsch, K.; Marcus, J.; Mossin, A.; and Dean, J. 2018.
\newblock Scalable and Accurate Deep Learning for Electronic Health Records.
\newblock \emph{npj Digital Medicine} 1(1).

\bibitem[{Ribeiro, Singh, and Guestrin(2016)}]{ribeiro16}
Ribeiro, M.~T.; Singh, S.; and Guestrin, C. 2016.
\newblock ``Why Should I Trust You?'' Explaining the Predictions of Any
  Classifier.
\newblock In \emph{ACM International Conference on Knowledge Discovery and Data
  Mining, SIGKDD}.

\bibitem[{Sakoe and Chiba(1978)}]{sakoe78}
Sakoe, H.; and Chiba, S. 1978.
\newblock Dynamic Programming Algorithm Optimization for Spoken Word
  Recognition.
\newblock \emph{IEEE Transactions on Acoustics, Speech, and Signal Processing}
  26(1).

\bibitem[{Sch{\"{a}}fer(2016)}]{Schafer16Scalable}
Sch{\"{a}}fer, P. 2016.
\newblock Scalable Time Series Classification.
\newblock \emph{Data Mining and Knowledge Discovery} 30(5).

\bibitem[{Schäfer and Leser(2017)}]{schafer17}
Schäfer, P.; and Leser, U. 2017.
\newblock Fast and Accurate Time Series Classification with WEASEL.
\newblock In \emph{ACM International Conference on Information and Knowledge
  Management, CIKM}.

\bibitem[{Senin and Malinchik(2013)}]{senin13}
Senin, P.; and Malinchik, S. 2013.
\newblock SAX-VSM: Interpretable Time Series Classification Using SAX and
  Vector Space Model.
\newblock In \emph{IEEE International Conference on Data Mining, ICDM}.

\bibitem[{Sorensen et~al.(1987)Sorensen, Jones, Heideman, and
  Burrus}]{Sorensen87realDFT}
Sorensen, H.; Jones, D.; Heideman, M.; and Burrus, C. 1987.
\newblock Real-valued Fast Fourier Transform Algorithm.
\newblock \emph{IEEE Transactions on Acoustics, Speech, and Signal Processing}
  35(6).

\bibitem[{Susto, Cenedese, and Terzi(2018)}]{Susto18cyber}
Susto, G.~A.; Cenedese, A.; and Terzi, M. 2018.
\newblock Time-Series Classification Methods: Review and Applications to Power
  Systems Data.
\newblock In Arghandeh, R.; and Zhou, Y., eds., \emph{Big Data Application in
  Power Systems}, chapter~9, 179--220. Elsevier.

\bibitem[{Tavenard et~al.(2017)Tavenard, Faouzi, Vandewiele, Divo, Androz,
  Holtz, Payne, Yurchak, Ru{\ss}wurm, Kolar, and Woods}]{tavenard17}
Tavenard, R.; Faouzi, J.; Vandewiele, G.; Divo, F.; Androz, G.; Holtz, C.;
  Payne, M.; Yurchak, R.; Ru{\ss}wurm, M.; Kolar, K.; and Woods, E. 2017.
\newblock tslearn: A Machine Learning Toolkit for Time Series Data.

\end{thebibliography}

\end{document}